\documentclass[twoside,twocolumn,9pt]{article}
\usepackage{extsizes}
\usepackage[super,sort&compress,comma]{natbib} 
\usepackage[version=3]{mhchem}
\usepackage[left=1.5cm, right=1.5cm, top=1.785cm, bottom=2.0cm]{geometry}
\usepackage{balance}
\usepackage{mathptmx}
\usepackage{sectsty}
\usepackage{graphicx}
\usepackage{subcaption}
\usepackage{booktabs}%
\usepackage{lastpage}
\usepackage{comment}
\usepackage[format=plain,justification=justified,singlelinecheck=false,font={stretch=1.125,small,sf},labelfont=bf,labelsep=space]{caption}
\usepackage{float}
\usepackage{fancyhdr}
\usepackage{fnpos}
\usepackage[english]{babel}
\usepackage{array}
\usepackage{droidsans}
\usepackage{charter}
\usepackage[T1]{fontenc}
\usepackage[usenames,dvipsnames]{xcolor}
\usepackage{setspace}
\usepackage[compact]{titlesec}
\usepackage{hyperref}
\usepackage{amssymb}

\usepackage{epstopdf}

\definecolor{cream}{RGB}{222,217,201}

\begin{document}

\pagestyle{fancy}
\thispagestyle{plain}
\fancypagestyle{plain}{
\renewcommand{\headrulewidth}{0pt}
}

\makeFNbottom
\makeatletter
\renewcommand\LARGE{\@setfontsize\LARGE{15pt}{17}}
\renewcommand\Large{\@setfontsize\Large{12pt}{14}}
\renewcommand\large{\@setfontsize\large{10pt}{12}}
\renewcommand\footnotesize{\@setfontsize\footnotesize{7pt}{10}}
\makeatother

\renewcommand{\thefootnote}{\fnsymbol{footnote}}
\renewcommand\footnoterule{\vspace*{1pt}%
\color{cream}\hrule width 3.5in height 0.4pt \color{black}\vspace*{5pt}} 
\setcounter{secnumdepth}{5}

\makeatletter 
\renewcommand\@biblabel[1]{#1}            
\renewcommand\@makefntext[1]%
{\noindent\makebox[0pt][r]{\@thefnmark\,}#1}
\makeatother 
\renewcommand{\figurename}{\small{Fig.}~}
\sectionfont{\sffamily\Large}
\subsectionfont{\normalsize}
\subsubsectionfont{\bf}
\setstretch{1.125} 
\setlength{\skip\footins}{0.8cm}
\setlength{\footnotesep}{0.25cm}
\setlength{\jot}{10pt}
\titlespacing*{\section}{0pt}{4pt}{4pt}
\titlespacing*{\subsection}{0pt}{15pt}{1pt}

\fancyfoot{}
\fancyfoot[R]{\footnotesize{\thepage}}
\fancyhead{}
\renewcommand{\headrulewidth}{0pt} 
\renewcommand{\footrulewidth}{0pt}
\setlength{\arrayrulewidth}{1pt}
\setlength{\columnsep}{6.5mm}
\setlength\bibsep{1pt}

\makeatletter 
\newlength{\figrulesep} 
\setlength{\figrulesep}{0.5\textfloatsep} 

\newcommand{\topfigrule}{\vspace*{-1pt}%
\noindent{\color{cream}\rule[-\figrulesep]{\columnwidth}{1.5pt}} }

\newcommand{\botfigrule}{\vspace*{-2pt}%
\noindent{\color{cream}\rule[\figrulesep]{\columnwidth}{1.5pt}} }

\newcommand{\dblfigrule}{\vspace*{-1pt}%
\noindent{\color{cream}\rule[-\figrulesep]{\textwidth}{1.5pt}} }

\makeatother

\twocolumn[
  \begin{@twocolumnfalse}
\begin{center}

\noindent\LARGE{\textbf{Quantum enhanced ensemble GANs for anomaly detection in continuous biomanufacturing$^\dag$}}
\vspace{1em}

\noindent\large{Rajiv Kailasanathan\textit{$^{a,d}$}, William R. Clements\textit{$^{b}$}, Mohammad Reza Boskabadi\textit{$^{a}$}, Shawn M. Gibford\textit{$^{a}$}, Emmanouil Papadakis\textit{$^{c}$}, Christopher J. Savoie\textit{$^{e}$}\ and Seyed Soheil Mansouri$^{\ast}$\textit{$^{a}$}}
\vspace{1em}
\end{center}

\noindent\normalsize{The development of continuous biomanufacturing processes requires robust and early anomaly detection, since even minor deviations can compromise yield and stability, leading to disruptions in scheduling, reduced weekly production, and diminished economic performance. These processes are inherently complex and exhibit non-linear dynamics with intricate relationships between process variables, thus making advanced methods for anomaly detection essential for efficient operation. In this work, we present a novel framework for unsupervised anomaly detection in continuous biomanufacturing based on an ensemble of generative adversarial networks (GANs). We first establish a benchmark dataset simulating both normal and anomalous operation regimes in a continuous process for the production of a small molecule. We then demonstrate the effectiveness of our GAN-based framework in detecting anomalies caused by sudden feedstock variability. Finally, we evaluate the impact of using a hybrid quantum/classical GAN approach with both a simulated quantum circuit and a real photonic quantum processor on anomaly detection performance. We find that the hybrid approach yields improved anomaly detection rates. Our work shows the potential of hybrid quantum/classical approaches for solving real-world problems in complex continuous biomanufacturing processes.}
\vspace{3em}

\end{@twocolumnfalse}
  ]

\renewcommand*\rmdefault{bch}\normalfont\upshape
\rmfamily
\section*{}
\vspace{-1cm}

\footnotetext{\textit{$^{a}$~Department of Chemical and Biochemical Engineering, Technical University of Denmark, Kongens Lyngby, Denmark}}
\footnotetext{\textit{$^{b}$~ORCA Computing, London, UK}}
\footnotetext{\textit{$^{c}$~Novo Nordisk, Kalundborg, Denmark}}
\footnotetext{\textit{$^{d}$~Novo Nordisk Foundation Centre for Biosustainability, Technical University of Denmark, Kongens Lyngby, Denmark}}
\footnotetext{\textit{$^{e}$~SiC Systems Inc., Franklin, TN, USA}}
\footnotetext{\textit{These authors have contributed equally: Rajiv Kailasanatha, William Clements and Mohammad Reza Boskabadi}}
\footnotetext{\textit{*Corresponding author: Seyed Soheil Mansouri (seso@kt.dtu.dk)}}
\footnotetext{\dag~Data and code available: https://github.com/RajivKailasanathan/ADCoB}

\section{Introduction}

The global biomanufacturing industry is expected to be valued between 500B\$ - 1T\$ by 2035\cite{thomas_measuring_2024}, and one of the key technological advancements required to achieve sustainable bioprocesses is the shift from batch to continuous mode of operation. Several challenges lie in the way of developing continuous processes, some of which are specific to bioprocesses, such as variation in quality of feedstock, genetic instability after many generations, or contamination risk. Given the high cost of operation of these processes, it is necessary to detect any anomalies in the process at a very early stage in order to intervene and prevent loss. This is also further reinforced in the form of strict regulatory requirements by several agencies that place strict regulations not only on the final product quality, but also on the quality of the process\cite{su_perspective_2019}. These requirements motivate the need for robust anomaly detection tools that can inform the operator of any anomalies in the process at an early stage. 

Through steady advancements of process monitoring tools and the growing availability of high-resolution data, industries across domains, including biomanufacturing, are making significant strides in the areas of model based process understanding and the development of digital twins\cite{lattanzi_digital_2021, boskabadi2025industrial, hernandez2025hybrid}. The development of robust methods for handling multivariate process data with multiple features is an active area of scientific and engineering research. One of the most important requirements of these methods is to detect anomalies in the process data, ideally as early as possible in the process, so that the process operator can make appropriate decisions about the changes in the process to prevent error and loss of yield or entire batches \cite{boskabadi2025virtual}.   

Deep learning-based approaches have been explored for anomaly detection in other industrial processes. Particularly, unsupervised methods that attempt to learn the characteristics of normal operational data to generate realistic synthetic process data have been explored for anomaly detection in a wide variety of fields. These include medicine\cite{lim_doping_2018,schlegl2019f}, video surveillance\cite{nguyen_anomaly_2020}, and quality inspection\cite{lai_industrial_2018}. In the field of industrial process monitoring, generative models, particularly Generative Adversarial Networks (GANs), for anomaly detection are being explored in various sectors including nuclear and conventional power plant operation \cite{choi_gan-based_2020,jintao_data_2019,li_research_2022}, steam turbine health\cite{que_semi-supervised_2019}, Oil and gas production\cite{du_multi-dimensional_2024}, fused magnesium furnace\cite{lu_deep_2019}. Anomaly detection strategies based on generative modeling tackle a critical and practical challenge: imbalanced process data. In industrial processes, normal operation is usually ensured through strict operational practices and as a result anomalous operation is usually rare. Hence, there is a huge disparity between the data that is available between normal operation and anomalous operation\cite{kim_gan-based_2020}. The modeling strategy for anomaly detection thus must be agnostic of the size of the anomalous process dataset. Generative models adopt an unsupervised learning strategy, and thus can be developed using only normal operational data. 

However, one of the biggest obstacles faced by academic researchers is the availability of industrial process data~\cite{mansouri2025models}. To overcome this challenge, simulation models of the real physical processes are used to create data whenever possible. Particularly in chemical processes, the Tennessee Eastman simulated process dataset~\cite{downs_plant-wide_1993} has been widely used to benchmark emerging methods in anomaly detection in industrial processes, particularly using generative models~\cite{yang_generative_2019,pozdnyakov_adversarial_2024,schoch_deep_2024,hartung_deep_2023}. 

Meanwhile, quantum GANs (QGANs) have emerged at the intersection of quantum computing and generative modeling. In these architectures, a quantum circuit makes use of the fundamental principles of quantum mechanics, including superposition, interference, and entanglement, to generate samples from probability distributions that are intractable to simulate classically~\cite{du_expressive_2020, andersson2022quantum}. Recent studies combine classical generative models with parameterized quantum circuits to create hybrid quantum-classical algorithms and observe improved performance~\cite{jin_improving_2025, kao_exploring_2023, li_quantum_2021, rudolph_generation_2022, wilson_quantum-assisted_2021}. Though the mechanisms that lead to this improved performance are still being elucidated, it has been suggested that the highly correlated and non-separable nature of quantum distributions can help neural networks produce more diverse data~\cite{bacarreza2025quantum}. In the context of industrial process monitoring, hybrid quantum classical algorithms are thus an attractive strategy for advancing anomaly detection methods. Early applications of hybrid quantum-classical methods for anomaly detection have shown promise~\cite{herr_anomaly_2021, kalfon_successive_2024, ajagekar_quantum_2020}.  

In this work, we develop a framework for anomaly detection in a continous biomanufacturing process. Biomanufacturing is a complex process with multiple sequential unit operations, feedback loops, and control units that results in highly non-linear process dynamics and complex relationships between process variables. To generate process data, we employ KTB-1, an established dynamic simulation model of a continous process for production of Lovastatin~\cite{boskabadi_kt-biologics_2024}, a commercially relevant Active Pharmaceutical Ingredient. Leveraging insight from domain knowledge, we also introduce faults in the simulation that result in process anomalies. Thus, we establish a benchmark dataset for normal and faulty operations of the continous biomanufacturing process. Subsequently, we investigate the applicability of GANs for anomaly detection in a continuous biomanufacturing process. We also extend our study to explore the potential of complex latent distributions generated by a quantum simulator and a real photonic quantum circuit in enhancing the performance of the GANs.

The key contributions of this paper are:
\begin{enumerate}
    \item Development of a benchmark dataset for normal and faulty operations of a continous biomanufacturing process. Sudden variation in the quality of feedstock is modelled as a fault, which aligns closely with real process deviations that are unique to bioprocesses.
    \item Development of an early anomaly detection framework using an ensemble of GANs. While GANs have applied for anomaly detection in industrial processes, this is the first study employing GANs for anomaly detection in continuous biomanufacturing.
    \item Novel application of quantum computers to enhance the performance of GANs. We explore the potential of hybrid quantum-classical algorithms in advancing the development of new processes in biomanufacturing.
\end{enumerate}

To ensure transparency and reproducibility, we also make our dataset and code publicly available at \url{https://github.com/RajivKailasanathan/ADCoB}.

\section{Methodology}

\subsection{Overview}

\begin{figure}[htp]
        \centering
        \includegraphics[width=\linewidth]{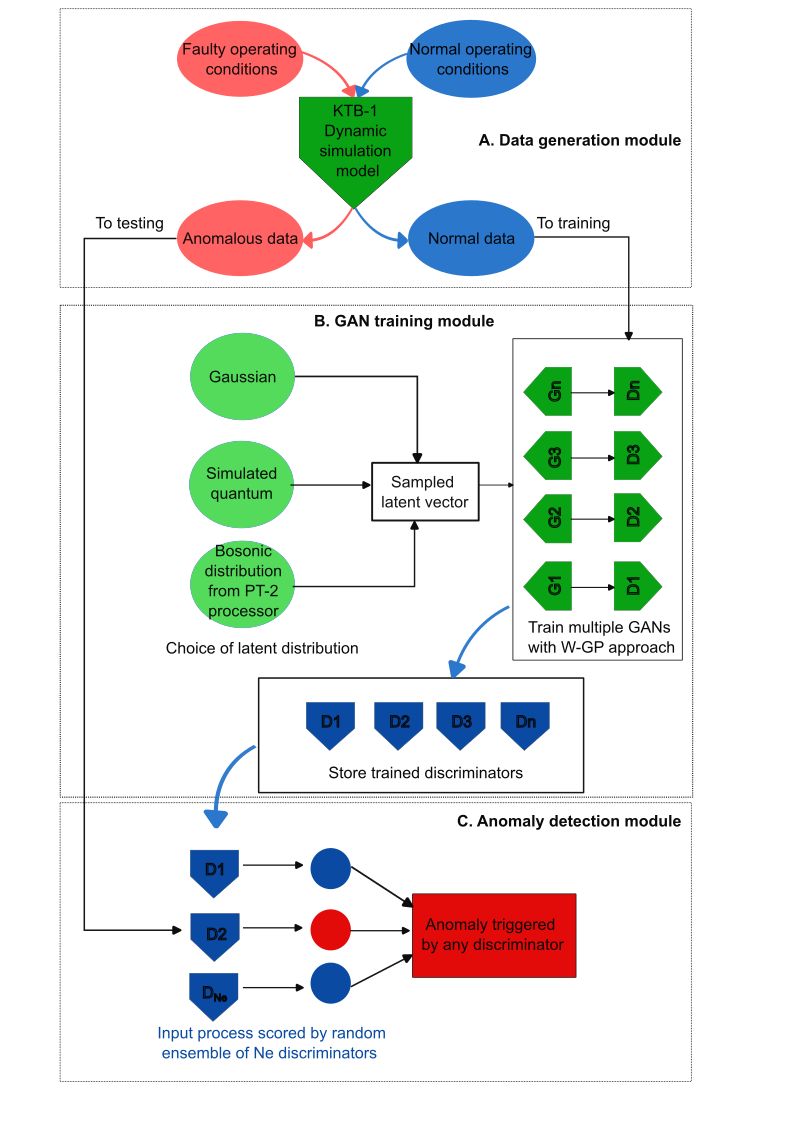}
    \caption{The anomaly detection workflow using an ensemble of GANs proposed in this study. \textbf{A}: first, using the KTB1 model for simulating the production process for lovastatin, a pharmaceutical ingredient, we produce a set of time series representing both normal and anomalous operating regimes. \textbf{B}: we train an ensemble of generative adversarial networks (GANs), each consisting of a generator ($G$) and discriminator ($D$) neural network, on the data from the normal regime. The generator is trained to map samples from an input probability distribution to synthetic data. Our work compares GANs trained with three different types of distributions: a classical Gaussian distribution, a simulated quantum distribution, and a real quantum distribution produced by an ORCA Computing PT-2 system. \textbf{C}: During the training process, each discriminator learns a decision boundary distinguishing between the normal operation data and other data. An ensemble of discriminators can thus be used as a detection anomaly module providing a set of learned criteria for flagging anomalies. If any discriminator in the ensemble flags a time series as anomalous, then the anomaly detection module labels the time series as an anomaly.}
    \label{fig:workflow}
\end{figure}

The anomaly detection workflow proposed in this study consists of a decomposed solution strategy that is applicable for any industrial time-series process data set. Figure~\ref{fig:workflow} shows an overview of the anomaly detection framework.

We use an unsupervised approach to train an ensemble of generative adversarial networks (GANs). This training section uses only normal operational data of the process, and does not require any anomalous data. As a case study, we demonstrate the use of this framework on a continous biomanufacturing process. A dynamic simulation model of the process (KT Biologics-1) is used to generate data required to train and test the model. By providing standard operation conditions (including normal fluctuations in process parameters), 'normal' operational data are generated. Similarly, sudden faults in the operational conditions are introduced to create anomalous data. The normal operation data is exclusively used to train the GANs. We also investigate different variations of a GAN, where we change the type of input distribution used by the generator to seed the synthetic data generation process. We compare the performance of three different distributions: a Gaussian distribution, a distribution generated by a simulated quantum circuit, and a distribution generated by a real photonic processor (ORCA Computing PT-2).

After the training process, the discriminators of the GANs are stored and used in the anomaly detection module. The input process is classified as anomalous if any one of the discriminators detects that it lies out of the distribution of the normal (training) data. 

\subsection{Dataset generation}

Our work uses the KTB1 simulation model, implemented in MATLAB Simulink, to generate synthetic anomalies across a plantwide biomanufacturing system. The simulation provides a dynamic and modular environment for modeling process behavior and fault conditions, which has, for example been used in prior work evaluating machine learning approaches for parameter estimation \cite{shahhoseyni_hybrid_2025}. The selected case study focuses on the production of lovastatin, an active pharmaceutical ingredient (API), using \textit{Aspergillus terreus} as the fungal host organism. The simulated process captures both upstream and downstream operations, replicating the integrated nature of industrial-scale biomanufacturing.

In the upstream section, raw materials, adenine and lactose, serving as nitrogen and carbon sources, are introduced into a mixing unit (M-101), where they are homogenized into a uniform feed solution. This medium is continuously supplied to a Continuous Stirred-Tank Reactor (CSTR), R-101, where Aspergillus terreus ferments the feedstock to biosynthesize lovastatin. The resulting fermentation broth is directed to a hydro-cyclone separator (HC-101), which separates the suspension into two output streams: a biomass-rich stream (R), part of which is recycled to the reactor to sustain biomass levels, and a lean stream (stream 5), which advances to downstream purification. This recycling loop introduces nonlinear dynamic behavior, making the process sensitive to disturbances and faults, conditions ideal for testing anomaly detection performance. 

\begin{figure*}[ht]
    \centering
        \includegraphics[width=\linewidth]{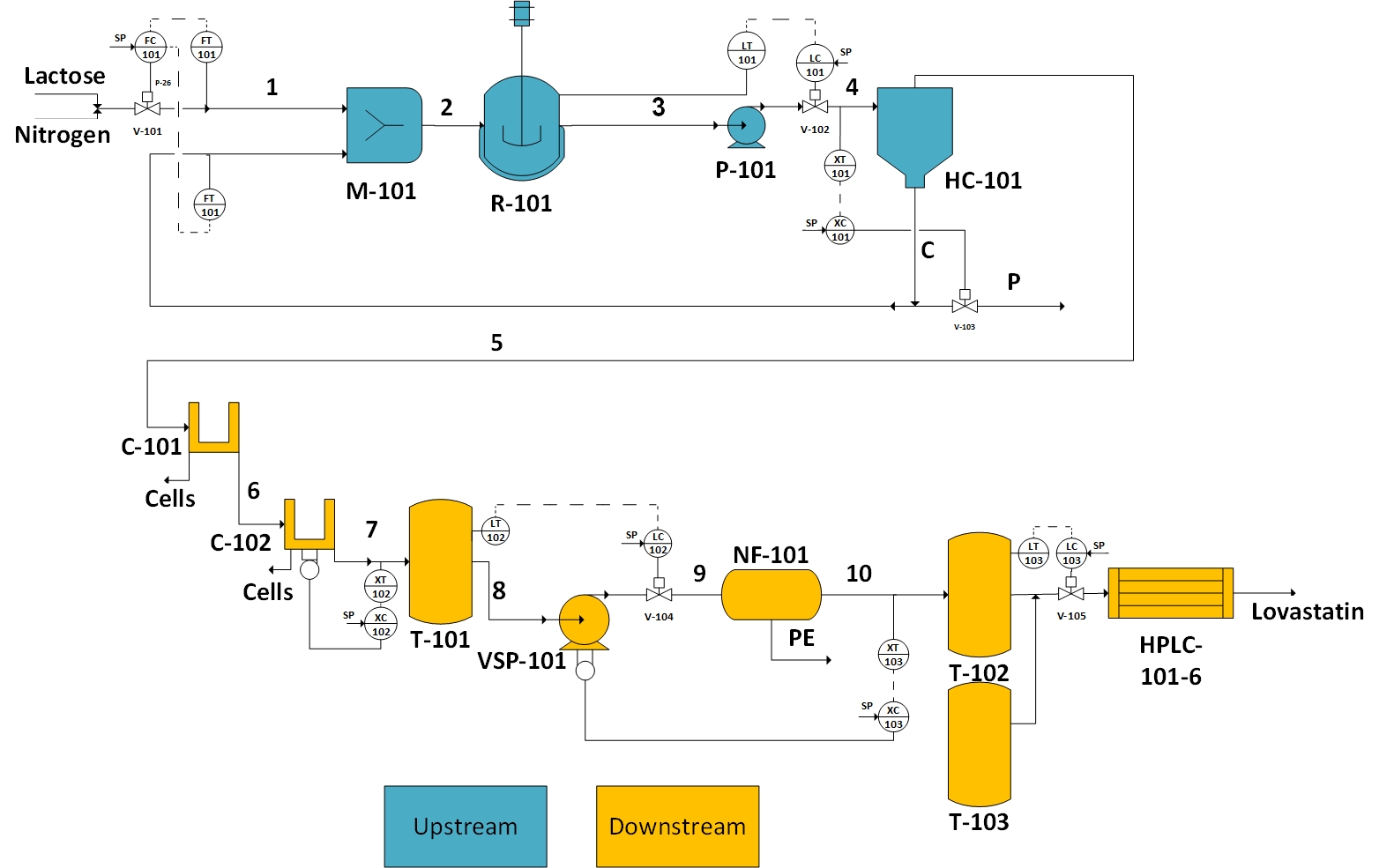}
    \caption{Process flow diagram of the KTB-1 simulation model. The upstream section focuses on enhancing Lovastatin production (with V for valves, R for reactors, P for pumps, and HC for hydrocyclones), while the downstream section is dedicated to Lovastatin purification and includes the associated control system for dynamic process management (where C stands for centrifuge, T for tank, and NF for nanofiltration).}
    \label{fig:process}
\end{figure*}

The downstream section consists of a sequence of separation and purification stages. Centrifuges C-101 and C-102 remove remaining biomass and dense particulates. The clarified broth is then concentrated via nanofiltration (NF-101), which retains lovastatin molecules while allowing smaller solutes to pass through. Final purification is performed through a series of six reversed-phase chromatography columns (LC-101 to LC-106), which isolate and enrich lovastatin based on molecular interactions. This complete Simulink-based process model enables the controlled injection of faults, such as flow deviations, concentration changes, membrane fouling, or sensor drifts across multiple units, producing realistic anomaly scenarios that challenge conventional detection methods. Figure \ref{fig:process} presents a process flow diagram summarizing the full system configuration.

\subsubsection*{Anomaly Dataset Generation}	

To develop and test fault detection and diagnosis algorithms, both normal and faulty datasets were generated using the KTB1 plant-wide dynamic simulator. Based on process operations, we define three operational modes: normal operation mode, disturbance mode, and faulty mode. There are no single universal numeric limits for "normal operation," "disturbance," and "fault" in biomanufacturing for parameters such as substrate concentration. Regulators and industry use concepts like Normal Operating Range (NOR), Proven Acceptable Range (PAR), design space, disturbance, and excursion/Out-of-Specification (OOS), expecting manufacturers to justify numeric ranges based on experiments, risk assessments, and control capabilities. To consider a worst-case and challenging scenario, we modeled normal operation mode with no intentional changes, disturbance mode with a tiny 0.5 percent step change in lactose concentration (Clac,M) from 20 g/L to 19.9 g/L, and faulty mode with a 1.5 percent step change from 20 g/L to 19.7 g/L. These changes propagate through the process and affect downstream behavior, as concentration variations in feed streams are recognized disturbances in biomanufacturing, as demonstrated in studies of continuous mAb production \cite{schmidt2022process}. 

Each simulation run captures the dynamic response of the plant under different operating conditions using Monte Carlo simulations with randomized parameters. For normal operating conditions, no fault or disturbance was introduced. A total of 1000 Monte Carlo simulations were conducted by varying the mean of Clac,M, using a normal distribution (mean = 20, variance = 0.01). This variation was implemented via a random number generator block, providing natural process variability while maintaining normal operation. Each simulation was run for 200 hours, and the output data was sampled at each hour.

For anomalous operating conditions (encompassing both disturbance and faulty modes), anomalies were introduced by applying step changes in Clac,M at time step 10 of the simulation. For the disturbance mode, the value was shifted from an initial baseline of 20 g/L to a final value of 19.9 g/L; for the faulty mode, it was shifted to 19.7 g/L. To make these anomalies more realistic and reflective of actual industrial conditions, the step signals were combined with a random noise generator, which simulates natural variability in the process. We used normally distributed noise with a mean of 0 and a variance of 0.01. Our initial testing found that these parameters created a challenging anomaly detection dataset, where the anomalies could be visually identified by plotting the data over the entire time series but were harder to identify at shorter time scales—particularly for the subtle disturbance mode, which represents a near-boundary challenge for differentiation from normal variability. As a result, before time step 10, the signal fluctuates randomly around the nominal value of 20 g/L. After time step 10, the signal continues to fluctuate, but now around the new anomalous value (19.9 g/L for disturbances or 19.7 g/L for faults). This composite signal allows the simulation to reflect both significant operational changes and continuous environmental variability.

A total of 100 time series were generated under each anomalous condition (disturbance and faulty modes separately). These time series were used only for evaluating the performance of the anomaly detection algorithm trained on normal data. The inclusion of both disturbance and faulty modes enhances the dataset's representation of real-world scenarios, where subtle drifts or excursions may precede more severe faults, testing the framework's sensitivity to varying anomaly magnitudes.

Further information about both the normal an anomalous datasets can be found in appendix \ref{datasetanalysis}, including representative plots showing the time series evolution of five quantities of interest.

\subsection{GANs for anomaly detection}

\begin{figure}[htp]
    \centering
        \includegraphics[width=\linewidth]{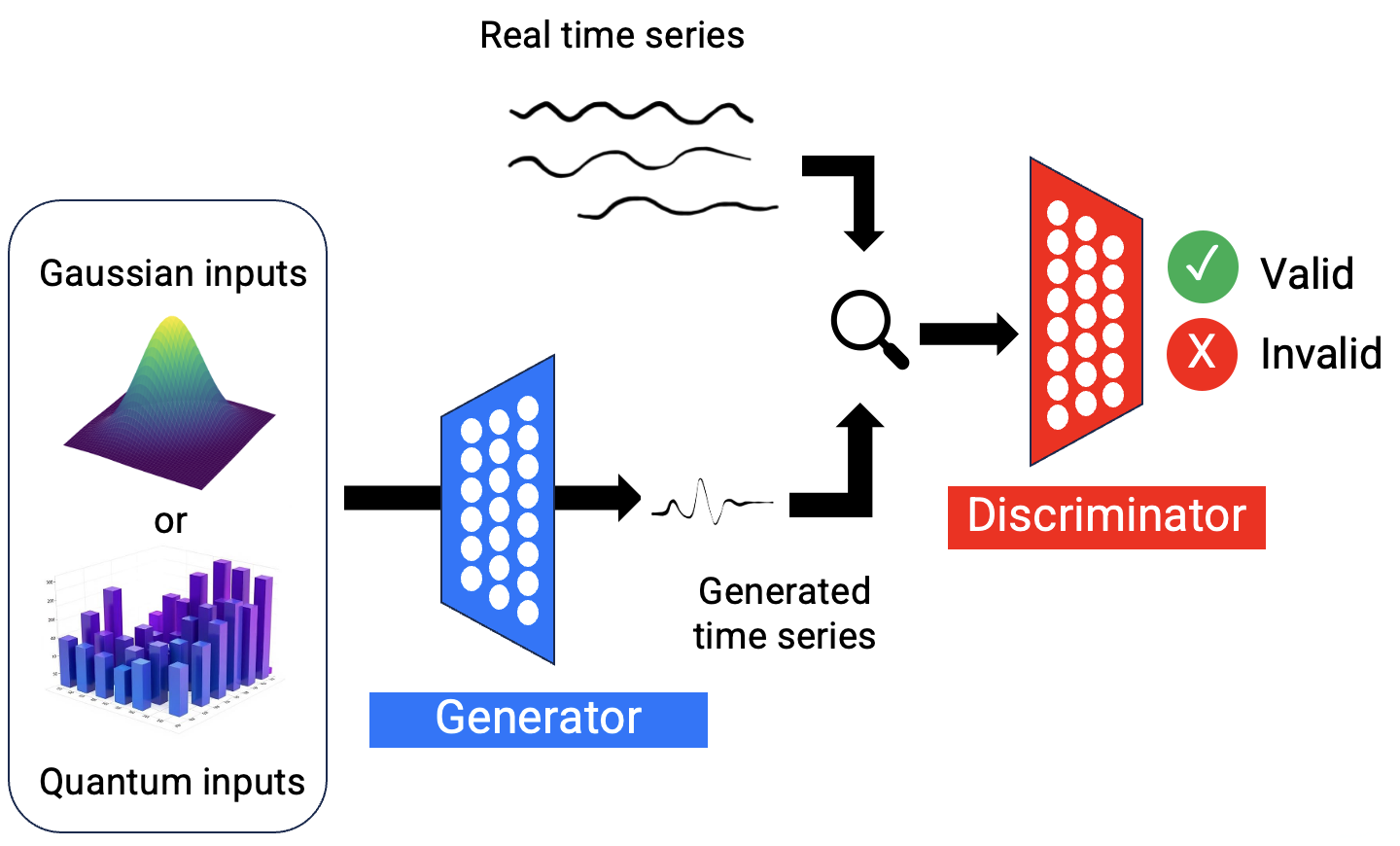}
    \caption{Our anomaly detection scheme uses generative adversarial networks, in which two neural networks are trained in an adversarial way. The generator is trained to map samples from an initial distribution to realistic synthetic data. In this work, we investigate using either a classical Gaussian distribution or a quantum distribution. The discriminator learns to distinguish between the real data and the synthetic data. After training, we discard the generator and use the discriminator to determine whether an input time series is anomalous.}
    \label{fig:gandiagram}
\end{figure}

We adopt an unsupervised anomaly detection method based on an ensemble of discriminator neural networks trained within a generative adversarial network (GAN) algorithm. Using GAN discriminators for unsupervised anomaly detection is a well-studied idea in machine learning \cite{schlegl2017unsupervised, schlegl2019f, akcay2018ganomaly, zenati2018adversarially}. For more robust anomaly detection, ensembles of discriminators have also been investigated \cite{han2021gan}. GAN-based methods for unsupervised anomaly detection have been proposed within a smart factory context \cite{kim2024novel}, but ours is the first to investigate this approach for bio-manufacturing.

As shown in figure \ref{fig:gandiagram} in a GAN a generator and a discriminator are trained in an adversarial manner. The discriminator is trained to distinguish between real data and synthetic data produced by the generator, and the generator is trained to produce synthetic data that can fool the discriminator \cite{goodfellow2014generative}. During training, the only real data that is used is the normal data, and no anomalous data is required. In this adversarial process, the discriminator effectively learns to distinguish between any out-of-distribution data produced by the generator and the normal data. After training, the generator is discarded and the discriminator is saved. When the discriminator is subsequently exposed to real anomalous data, the expectation is that it should flag it as such.

To make this method more robust, an ensemble of GANs can be used. A single discriminator learns the boundary between normal data and the specific out-of-distribution data produced by one generator. The discriminator thus learns one specific decision boundary in the problem space, but this is typically insufficient to detect a wide range of potential anomalies. Training multiple independent GANs can be an effective way of learning multiple such decision boundaries. Though training multiple GANs increases the computational cost of this method, in this work we consider lightweight models that can be trained in a reasonable amount of time.

In our specific implementation, we train an ensemble of Wasserstein GANs with gradient penalty \cite{gulrajani2017improved}. These are known to be more stable to train and yield higher-quality results than the original GAN proposal. Both the generator and the discriminator consist of feed-forward neural networks with 2 hidden layers and LeakyReLU activation. An important component of the generator is the "latent distribution". To produce a synthetic datapoint, the generator first samples a latent vector from the latent distribution, and this vector is used as the input to the neural network. The properties of this latent distribution are known to play an important role in the quality of the generated data \cite{hu2023complexity}. Our work considers size-16 latent vectors, which are drawn either from a Gaussian or a quantum probability distribution.

After training, we use a "union rule" to flag anomalies, where if any one of the ensemble of discriminators flags the data as anomalous then the data is flagged as anomalous. This reflects the notion that each discriminator learns a different decision boundary between normal data and out-of-distribution data, such that anomalous data will generally lie outside the decision boundary for some but not all models. Specifically, since each discriminator output is a real number, we first determine the bottom-$x$ and top-$x$ quantiles (for a given value of $x$) of this value for the normal data, then any data that is outside these thresholds is flagged as anomalous. Different values of $x$ lead to different true positive and false positive rates, and as such we present our results as receiver operating characteristic (ROC) curves. We also present the associated areas under the curve (AUC).

\subsection{Hybrid quantum/classical GANs}

Anomaly detection methods based on ensembles of GANs require using generator architectures that can produce highly diverse data. Producing highly diverse data is expected to help the discriminator learn more useful decision boundaries between normal data and out-of-distribution data within the problem space. However, despite significant improvements in GAN models since they were first proposed, GANs often struggle with "mode collapse", where data diversity can be minimal \cite{srivastava2017veegan}. 

Recently, several studies using quantum processors as part of a hybrid quantum/classical GAN architecture have found that these methods can outperform purely classical GANs \cite{rudolph_generation_2022,wilson_quantum-assisted_2021,li_quantum_2021,bacarreza2025quantum,xiao2024quantum}. Quantum processors can sample from probability distributions that are intractable classically \cite{hangleiter2022computational}. In these studies, samples drawn from these quantum distribution are used at the input of the generator instead of a classical distribution. This method has been observed to help the generator produce more diverse outputs \cite{xiao2024quantum,bacarreza2025quantum}. We hypothesized that generators that produce more diverse synthetic data should in turn produce discriminators that learn a more informative decision boundary between normal and out-of-distribution data. As such, in this study we investigate the performance of an ensemble of discriminators trained with hybrid quantum/classical generators, and compare it to the performance of an ensemble of discriminators trained using purely classical generators.

\begin{figure}[h]
    \centering
    \includegraphics[width=0.45\textwidth]{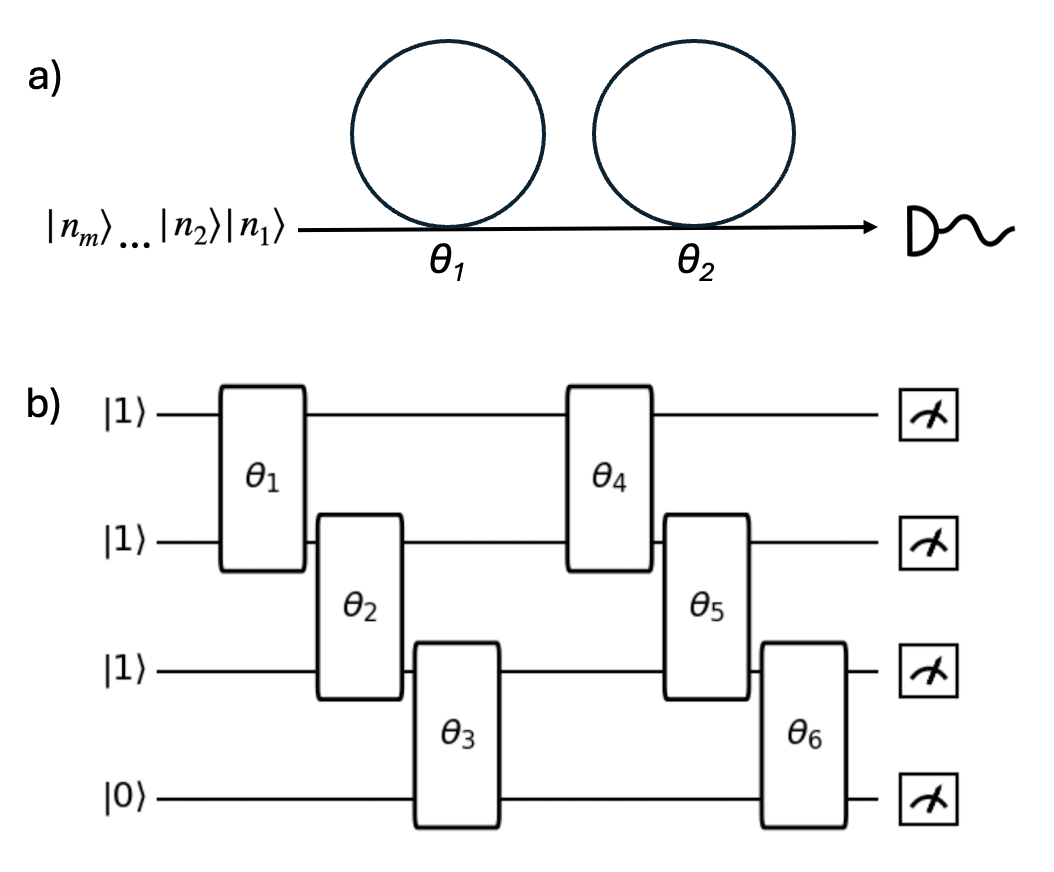}
    \caption{Our experiments using a quantum processor use photonic quantum processors implemented with a time-bin architecture. As shown in a, in these architectures single photons are sequentially sent into a network of optical delay lines with programmable coupling coefficients. This creates an entangled state between different time bins which can measured by a photon number resolving detector. The two-loop architecture shown in a) implements the quantum optical circuit shown in b), where each gate is a programmable beam splitter between two optical modes.}
    \label{fig:tbi}
\end{figure}

We use both simulated and real photonic quantum processors using the architecture shown in figure \ref{fig:tbi}. In a photonic quantum processor, identical single photons are injected into an interference network. Interference between the photons creates a complex quantum state, which is collapsed by a measurement process that returns the output locations of the photons. Simulating this process classically requires computational resources that scale exponentially with the number of photons, such that simulations with more than a few tens of photons are intractable \cite{aaronson2011computational}. Our experiments consider small-scale quantum processors with 8 photons interfering in 16 channels.

Photonic quantum processors have several advantages compared to other quantum computing modalities that motivate using them for hybrid machine learning algorithms. Though near-term photonic systems are not universal for quantum computation, they naturally produce highly non-uniform distributions \cite{aaronson2013bosonsampling} that differ significantly from commonly used classical latent distributions. Moreover, photonic systems are not subject to decoherence, which pushes the outputs of quantum circuits implemented with other modalities closer to a uniform distribution \cite{wang2021noise}. Finally, their room-temperature operation and commercial availability make their integration into real-world workflows more straightforward \cite{slysz2025hybrid}.

Our classical simulations of these processors use the algorithm from Clifford and Clifford \cite{clifford2018classical}, whereas our experiments with a real quantum processor use a commercially available ORCA Computing PT-2. We note that classical simulations of these systems are only tractable for the small-scale quantum systems considered in this work and become intractable at larger sizes. We use randomly initialized and untrained photonic quantum circuits to provide an apples-to-apples comparison with the untrained Gaussian distribution used for the classical GANs. However, we note that these circuits could in principle be trained jointly with the classical neural networks in the generator and discriminator \cite{facelli2024exact}.

\section{Results}

We perform experiments that aim to answer different research questions. First, we perform a range of experiments using ensembles of 30 models with different latent spaces. These experiments aim to validate our approach to anomaly detection, and elucidate how the choice and size of the latent space impact the overall model performance. Comparing the performance of the simulated quantum processor to that of the real quantum processor also allows us to examine the transferability of our method to real quantum hardware. Our second experiments use a larger ensemble of 60 models to investigate how model performance scales with the number of discriminators. 

\subsection{Experiments with an ensemble of 30 discriminators}

\begin{figure}[htp]
    \centering
        \includegraphics[width=\linewidth]{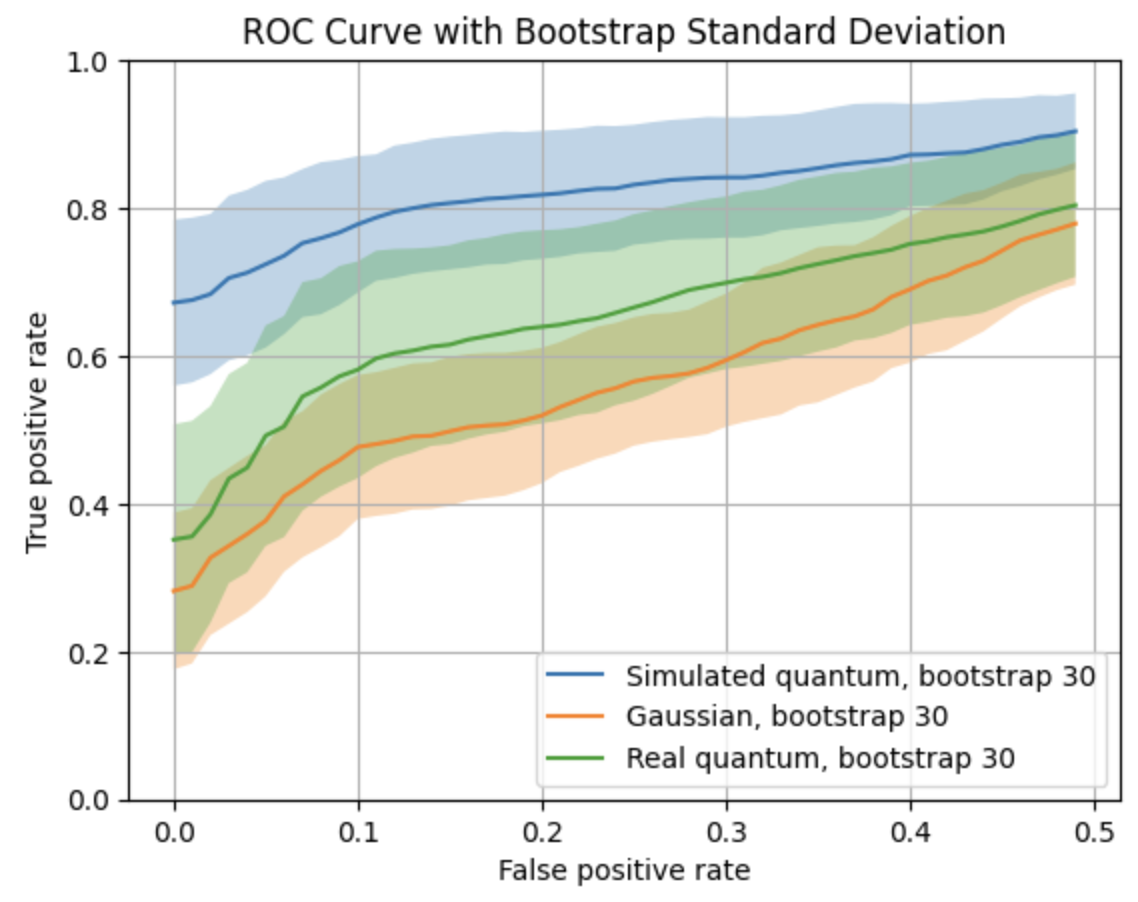}
    \caption{Receiver operating characteristic (ROC) curves of our anomaly detection algorithm based on an ensemble of 30 GAN discriminators. We consider GANs that use a classical Gaussian latent space or a quantum distribution produced by the quantum interference between identical photons. The quantum distribution is generated either by a real or by a simulated quantum processor. The GANs differ only in the process that produced the latent vectors sent to the generator. The solid line and shaded areas correspond respectively to the mean and standard deviation calculated by sampling 30 models 20 times from an ensemble of 120 trained models.}
    \label{fig:sim20}
\end{figure}

Our first experiments focus on using ensembles of 30 discriminators for anomaly detection. To evaluate the mean and standard deviation of the performance of an ensemble for anomaly detection, for each experiment we train 120 GAN models and use 20 bootstrap ensembles of 30 models drawn from this larger set.

\begin{table}[h!]
  \centering
  \caption{Area under the curve (AUC) of the ROC curve in figure \ref{fig:sim20}, with a size-30 bootstrap ensemble. Perfect detection = 0.5}\label{tab:auc}
  \begin{tabular}{lc}
    \toprule
    Latent space & AUC \\ 
    \midrule
    Simulated quantum distribution & 0.409 $\pm$ 0.012 \\
    Real quantum distribution & 0.324 $\pm$ 0.026 \\
    Gaussian distribution & 0.280 $\pm$ 0.023 \\
    \midrule
    \textit{Random guessing} & \textit{0.125} \\
    \bottomrule
  \end{tabular}
\end{table}

ROC curves from these different model ensembles are shown in figure \ref{fig:sim20}, with areas under the curve (AUC) shown in table \ref{tab:auc}. We observe that all approaches successfully detect anomalies, with some significant differences in performance between the models. These results validate the use of an ensemble of GAN discriminators as a promising method for unsupervised anomaly detection in a high-dimensional bio-manufacturing environment. 

We also observe that the best results, with over 50\% anomaly detection success at 0\% false positives, are achieved with the quantum distributions. These results indicate that the greater diversity of the data produced by the hybrid quantum/classical discriminators helped the discriminators learn more useful decision boundaries for anomaly detection. We also find that the real quantum processor outperforms the classical Gaussian distribution but underperforms the simulated quantum distribution. This suggests that experimental imperfections such as photon loss have a material impact on performance, but using even an imperfect real quantum processor can still yield better results than the most commonly used Gaussian distribution. These results validate the use of a hybrid quantum/classical approach for detecting anomalies.
    
\subsection{Using a larger ensemble}

\begin{figure}[h]
    \centering
    \includegraphics[width=\linewidth]{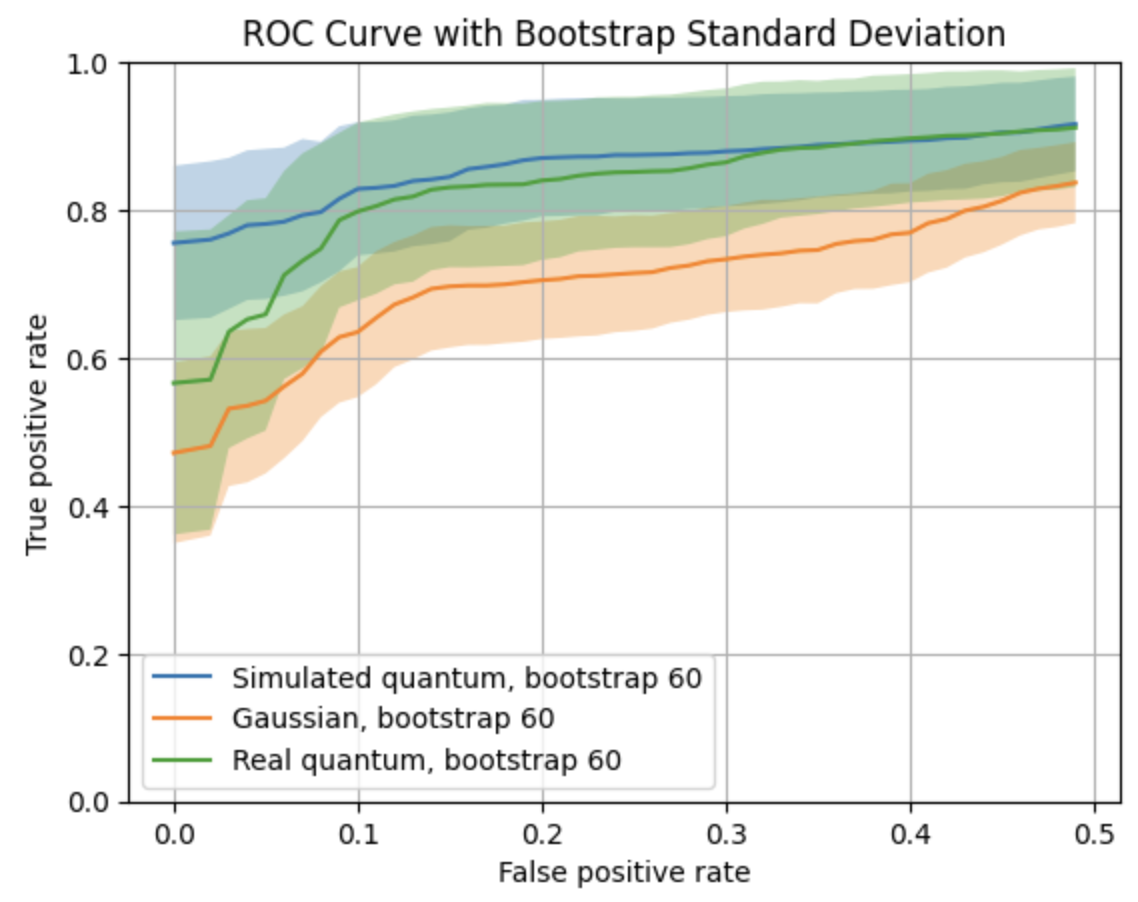}
    \caption{ROC curves of our anomaly detection method based on an ensemble of 60 GAN discriminators using different latent spaces.}
    \label{fig:aggregated}
\end{figure}

We also investigate the performance of our method using a larger ensemble of 60 models sampled from the size-120 ensemble, with results shown in figure \ref{fig:aggregated} and table \ref{tab:auc2}. We find that the larger ensemble yields improved results. Moreover, the hierarchy between models that was observed for the ensembles of 30 models is reflected here too. These results show that this method scales well with the amount of computation available. Whereas the use of the union rule method of flagging anomalies could have been expected to create a large number of false positives as the ensemble size increases, these results indicate that this is not a problem in practice.

\begin{table}[h!]
  \centering
  \caption{Area under the curve (AUC) of the ROC curve in figure \ref{fig:aggregated}, with a size-60 bootstrap ensemble. Perfect detection = 0.5}\label{tab:auc2}
  \begin{tabular}{lc}
    \toprule
    Latent space & AUC \\ 
    \midrule
    Simulated quantum distribution & 0.429 $\pm$ 0.009 \\
    Real quantum distribution & 0.412 $\pm$ 0.019 \\
    Gaussian distribution & 0.350 $\pm$ 0.019 \\
    \midrule
    \textit{Random guessing} & \textit{0.125} \\
    \bottomrule
  \end{tabular}
\end{table}

\section{Discussion}

\subsection{Strengths and weaknesses of a GAN-based approach}

Though our GAN-based approach is advantageous in terms of applicability to complex datasets that may arise in a biomanufacturing setting, some disadvantages must be acknowledged. First, using an ensemble of GANs can be computationally intensive, with size-30 and size-60 ensembles considered in our work. Though each model we consider is small and can be trained in under 5 minutes on an Apple M2 chip, scaling to larger models and datasets may be onerous. However, we note that after training, inference can often be performed quickly and cheaply. 

We also note that other deep learning based approaches have been developed for anomaly detection and could be applied here. Though GANs for time series anomaly detection such as BeatGAN \cite{zhou2019beatgan} have been successful, other approaches have been developed such as Anomaly Transformer \cite{xu2021anomaly}. However, our objective with this work was not to beat existing benchmarks for anomaly detection in biomanufacturing, but rather to develop and investigate a new approach for doing so. Further refining the method and comparing it to other methods on a broader range of problems would be a natural extension of this work.

\subsection{Advantage from quantum computing}

Here, we further discuss the role of quantum computing in helping us achieve improved anomaly detection results. Though this current work used a distribution that is native to quantum processors, the small-scale nature of the distribution with 8 photons in 16 qumodes makes it classically simulable. However, scaling this approach to larger-scale datasets is likely to require larger processors that are no longer simulable. For these larger-scale quantum processors, we consider two questions: whether the quantum computer is required to achieve higher-quality results, and the extent to which the quantum computer could be, if not exactly simulated, at least approximated classically. 

First, though the quantum distribution provides a practical and well-studied method to improve the performance of a GAN, it is not the only method available. A wide range of other classical distributions can also be used in a GAN, from Bernoulli distributions~\cite{brock2018large} to trained distributions using an autoencoder~\cite{hu2023complexity}. Though Bacarreza et al.~\cite{bacarreza2025quantum} observe an improvement over a wide range of classical distributions, we cannot rule out that some of these methods may yield improved performance on this problem compared to the baseline Gaussian. However, it is impractical and computationally expensive to investigate a wide range of alternative distributions, while the increasing availability of quantum computers means that they are becoming an increasingly practical alternative. As such, in realistic settings where algorithm development time and computation is limited, adopting a quantum distribution as the go-to option for improving performance may be a more effective use of resources than investigating a wider range of classical distributions.

Second, we note that classical methods for simulating photonic quantum processors in the presence of noise have been developed~\cite{villalonga2021efficient, oh2024classical}. The methods could in principle be used to derive the same improvement in performance that we observe with a real quantum processor using only classical resources. However, these methods are not practical for hybrid quantum/classical machine learning workflows. The mean-field approach of Villalonga et al.~\cite{villalonga2021efficient} reproduces the low-order statistics of a photonic quantum distribution with a quadratic overhead as opposed to a linear overhead at most with a real system. The approach of Oh et al.~\cite{oh2024classical} was reported to require 9 minutes on 288 GPUs just to set up the simulation. As such, it is unlikely that either approach could run faster than a state-of-the-art photonic quantum processor. Though speed of generating samples is less of a concern when using a static, untrained latent space, this can be a major limitation for more advanced variational settings. In those settings where both the quantum and the classical parts of the algorithm are regularly updated, neither approach thus provides a practical route to simulating these distributions.

\subsection{Prospects for practical biomanufacturing applications}
The convergence of quantum-enhanced machine learning and advanced control strategies presents a transformative opportunity for continuous biomanufacturing. As processes become increasingly complex and data-rich, traditional control systems struggle to maintain robustness in the face of nonlinear dynamics, rare anomalies, and evolving operational conditions. In this context, the integration of hybrid quantum Generative Adversarial Networks (GANs) with Model Predictive Control (MPC) offers a promising pathway toward intelligent, adaptive process management.

Hybrid quantum GANs leverage the expressive power of quantum-generated latent spaces to produce synthetic data that more accurately reflects the manifold of normal process behavior. This enhanced diversity enables discriminators to learn sharper decision boundaries, improving the sensitivity and specificity of anomaly detection. Importantly, these models operate in an unsupervised fashion, relying solely on normal operational data—an essential feature in industrial settings where labeled anomalies are scarce or unavailable.

Embedding this anomaly detection capability within an MPC framework creates a closed-loop system that not only monitors but also responds to deviations in real time. When the GAN ensemble flags an anomaly, the MPC module can dynamically adjust its prediction horizon, tighten constraints, or re-optimize control actions to mitigate potential risks. This dual-layered architecture—where anomaly detection informs control decisions—represents a shift from reactive to proactive process management.

Moreover, the inclusion of direct control signals from the anomaly detection module to the process simulation or plant adds a layer of safety and responsiveness. In scenarios where anomalies indicate critical faults, bypassing the MPC for immediate actuation can prevent cascading failures or product loss. This hybrid control logic, combining predictive optimization with anomaly-triggered overrides, aligns well with the stringent reliability and quality demands of pharmaceutical manufacturing.

From a broader perspective, this framework exemplifies how quantum computing can be pragmatically integrated into industrial workflows. Photonic quantum processors, with their room-temperature operation and commercial availability, offer a feasible route to enhancing machine learning models without requiring full-scale quantum infrastructure. As quantum hardware matures, its role in enabling more expressive generative models—and by extension, more intelligent control systems—is likely to expand.

In summary, the proposed Hybrid Quantum GAN-MPC framework is not merely a technical innovation but a conceptual advancement in how we think about process control. It bridges the gap between data-driven anomaly detection and model-based control, paving the way for resilient, adaptive, and intelligent biomanufacturing systems.

\section{Conclusions}

In this study, we have introduced a novel framework for unsupervised anomaly detection in continuous biomanufacturing processes, leveraging an ensemble of generative adversarial networks (GANs) enhanced by hybrid quantum-classical approaches. By developing a benchmark dataset using the KTB-1 dynamic simulation model for lovastatin production, we simulated both normal operations and anomalies arising from sudden feedstock variability—a common challenge in bioprocesses. Our ensemble GAN method, trained exclusively on normal data, demonstrated robust early detection capabilities, with receiver operating characteristic (ROC) curves indicating effective discrimination between normal and anomalous regimes.

A key innovation lies in the integration of quantum-generated latent distributions, sourced from both simulated and real photonic quantum processors (ORCA Computing PT-2). Comparative experiments revealed that these quantum distributions outperformed classical Gaussian baselines, yielding higher areas under the curve (AUC) values—up to 0.429 for simulated quantum ensembles of 60 discriminators. This improvement underscores the potential of quantum mechanics to enhance generative diversity, mitigating issues like mode collapse and enabling discriminators to learn more informative decision boundaries. Notably, even imperfect real quantum hardware provided advantages over classical methods, validating the practicality of hybrid architectures in near-term applications.

This work represents the first application of hybrid quantum-classical GANs for anomaly detection in continuous biomanufacturing, addressing critical needs for robust process monitoring amid nonlinear dynamics and regulatory demands. The framework's unsupervised nature and scalability with ensemble size position it as a promising tool for integration with model predictive control (MPC) systems, facilitating proactive interventions to minimize yield losses and operational disruptions.

While our results highlight quantum advantages, limitations include reliance on small-scale quantum processors and the computational cost of larger ensembles. Future research could explore trained quantum circuits, scalability to fault-tolerant quantum hardware, and real-world validation on industrial datasets. Overall, this study paves the way for quantum-enhanced machine learning to drive advancements in sustainable biomanufacturing, offering a pathway toward more resilient and efficient processes.

\section*{Author contributions}
\textbf{Rajiv Kailasanathan:} Methodology, Software, Investigation, Writing - Original Draft,  \textbf{William Clements:} Methodology, Software, Investigation, Writing - Original Draft, \textbf{Mohammad Reza Boskabadi:} Methodology, Investigation, Writing - Original Draft, \textbf{Shawn M. Gibford:} Software, Investigation, Writing - Review \& Editing, \textbf{Emmanouil Papadakis:} Writing - Review \& Editing, \textbf{Christopher J. Savoie:} Conceptualization, Methodology, Writing - Review \& Editing, \textbf{Seyed Soheil Mansouri} Conceptualization, Methodology, Supervision, Writing - Review \& Editing, Project administration, Funding acquisition

\section*{Conflicts of interest}
There are no conflicts to declare.

\section*{Data availability}

All codes and datasets used in this study are available according to FAIR principles: https://github.com/RajivKailasanathan/ADCoB

\section*{Acknowledgements}

This work was funded by the Novo Nordisk Foundation through grant number NNF20CC0035580, and the Technical University of Denmark.
\bibliography{references} 
\bibliographystyle{ieeetr}

\appendix

\section{Data pre-processing}

We pre-processed the data in the following way. First, though each time series in the original dataset was 200 hours long, our initial investigation found that using the entire dataset caused the anomalies to be too easy to detect, with near-perfect anomaly detection performance that provides little insight into the effectiveness of our method. We thus focused on the early anomaly detection setting and limited the dataset to the first 48 hours (with the anomaly starting at the 10th hour). Since changes to the measured quantities are slow, we further downselected to only use data collected every 8th hour. We tracked 5 features in the time series that were found to be affected by the anomaly, such that each time series was reduced down to $(48/8) \times 5 = 30$ features after pre-processing. To help process this data with a neural network, we then standardized these features using the mean and standard deviation over time of each of the 5 quantities.

\section{Individual ROC curves}\label{nobootstrap}

\begin{figure}[h]
    \centering
    \includegraphics[width=\linewidth]{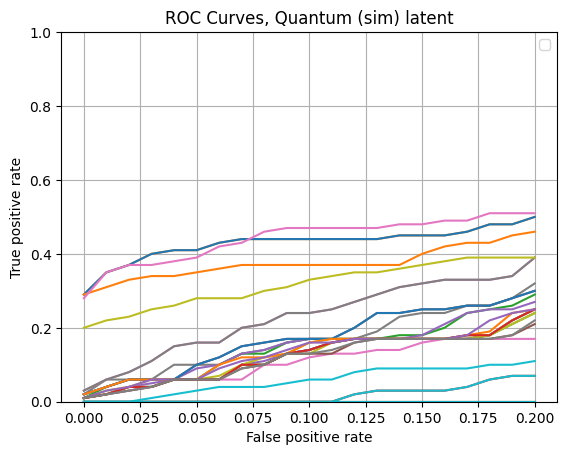}
    \includegraphics[width=\linewidth]{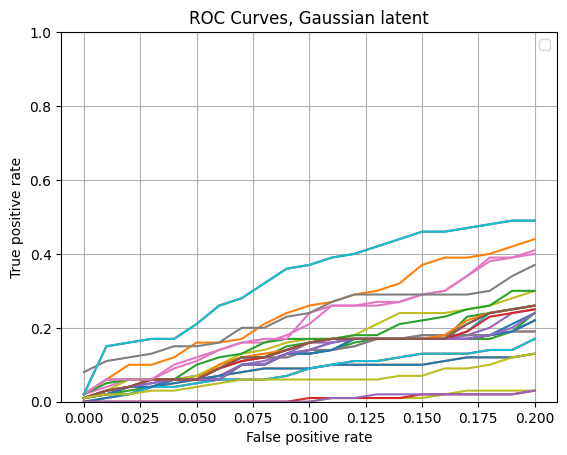}
    \caption{ROC curves of 30 different discriminators used on their own to detect anomalies for (top) simulated quantum and (bottom) Gaussian latent distributions.}
    \label{fig:singles}
\end{figure}

To provide insight into the behavior of individual constituents of the GAN ensembles used for anomaly detection, figure \ref{fig:singles} shows ROC curves for 30 randomly selected discriminators for two different latent distributions. First, we observe that every discriminator significantly underperforms the ensemble with the ROC curve in figure \ref{fig:sim20}. This shows that the ensembling provides a benefit. Second, we see very heterogeneous performance between different discriminators. Since all the GANs successfully trained to convergence, this suggests that different decision boundaries are indeed learned by the discriminators, and only some of which are useful for the type of anomaly investigates in this work. Third, we see more diversity in discriminator performance with the quantum latent distribution especially at lower false positive rates, with 4 discriminators significantly outperforming the others. This suggests that the quantum latent distribution learned a broader set of decision boundaries, leading to improved performance for a subset of the discriminators on this task.

\section{ORCA Computing PT-2 system}\label{pt2}

The photonic quantum processor used in this work is an ORCA Computing PT-2 system. The PT-2 consists of a time-bin interferometer as shown in figure \ref{fig:tbi}, where a single pulsed quantum light source produces a sequence of regularly spaced photons. These photons interfere with each other inside a sequence of two optical delay lines, where the length of each delay line matches the time delay between photons. This allows the photons to interfere with other photons in subsequent pulses, creating a complex superposition state. A single detector at the output can resolve up to 6 photons in each time bin.

The system we use in this work is hosted at the UK National Quantum Computing Centre. Our experiments use 16 time bin bins, with photons injected at each time bin. Due to system losses of roughly 70\%, fewer photons are detected than are sent in. To approximate the ideal simulation where 8 photons are detected, we use "post-selection", where we only return outcomes where at least 8 photons were detected. Other sources of imperfection, such as photon distinguishability, beam splitter inaccuracies or detector dark counts were all found to play a minor role compared to system loss.

For our experiments with a simulated quantum processor, we use the simulation algorithm by Clifford and Clifford \cite{clifford2018classical}. We consider an input state with single photons alternating with vacuum in 16 channels, for a total of 8 photons. These simulations assume an ideal quantum processor with no loss and perfect photon number resolution. The circuit we use is the one shown in figure \ref{fig:tbi}, with randomly selected beam splitter parameters. 

\section{Dataset analysis}\label{datasetanalysis}

Here, we provide additional information concerning the dataset used for this task. All data was generated from the same plant-wide dynamic simulator with identical process noise (Gaussian, $\sigma = 0.1$). Figure \ref{fig:features} shows typical trajectories of the 5 features we consider in the data for both the normal data used for training and anomalies in which a 0.5\% drop in lactose feed concentration occurred after 10 hours. We observe that for each feature, the anomalous data overlaps significantly with the normal data, with divergence only towards the end of the trajectory. Moreover, the gap between anomalous and normal data at the end of the trajectory is relatively small, with overlap in some the features.

\begin{figure}
    \centering
    \includegraphics[width=0.8\linewidth]{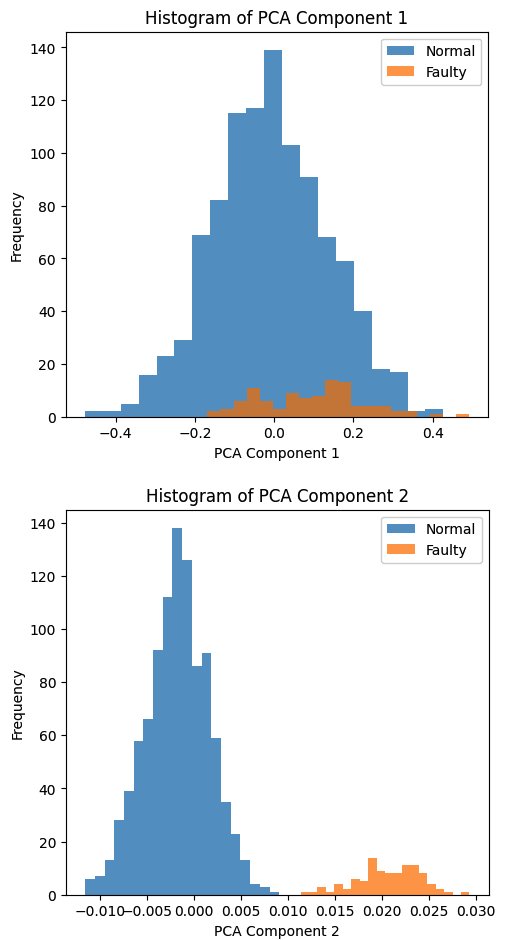}
    \caption{Histograms of the first (top) and second (bottom) PCA components of the normal (blue) and anomalous (red) data.
 }
        \label{fig:pca}
\end{figure}

To further quantify the difference between normal and anomalous data, in figure \ref{fig:pca} we plot histograms of both types of data along the first two PCA components of the normal data. We observe almost complete overlap in the first PCA component, which captures 95\% of the variance in the normal data. A difference between normal and anomalous data only becomes apparent when considering the second PCA component. These results illustrate the importance of learning good decision boundaries in the data. A simple model that learns a decision boundary based on the first PCA component will not identify these anomalies. A more refined model that is able to learn more complex decision boundaries is required. Though in this case a linear model based on the second PCA component in the data would have been successful, in the unsupervised anomaly detection scenario this is not information that would have been available a priori. Using a neural network for this task provides the flexibility to autonomously learn linear or non-linear decision boundaries where appropriate.

In addition to our experiments with a 0.5\% disturbance, we also initially investigated a dataset with a 1.5\% disturbance. This was however found to be too simple, leading us to focus our efforts instead on the 0.5\% dataset.

 \begin{figure}
     \centering
     \includegraphics[width=\linewidth]{figures/trajectories.jpg}
     \caption{Trajectories of the selected features. These correspond to some of the most important propagating control variables and measured variables, and are respectively recycle factor, product concentration after the nanofilter, recycle rate, product concentration after reactor, and lactose concentration after reactor. The blue curves are sample trajectories from the normal dataset, and the red curves are sample trajectories from the anomalous dataset.}
     \label{fig:features}
 \end{figure}

\balance

\end{document}